%% file: main.tex
\documentclass{article}

\usepackage{iclr2027_conference,times}

\iclrfinalcopy

\usepackage[utf8]{inputenc}
\usepackage[T1]{fontenc}
\usepackage{hyperref}
\definecolor{mydarkblue}{RGB}{0,70,140}
\hypersetup{
  colorlinks=true,
  citecolor=blue,
  linkcolor=black,
  urlcolor=mydarkblue
}
\usepackage{url}
\usepackage{adjustbox}
\usepackage{booktabs}
\usepackage{amsfonts}
\usepackage{amsmath,amssymb}
\usepackage{nicefrac}
\usepackage{microtype}
\usepackage[dvipsnames]{xcolor}
\definecolor{scacircuit}{HTML}{3B7D23}
\definecolor{scacomplement}{HTML}{C65F10}
\usepackage{colortbl}
\usepackage{tabularx}
\usepackage{multirow}
\usepackage{enumitem}
\usepackage{tcolorbox}
\usepackage[font=small]{caption}
\usepackage{placeins}
\usepackage{graphicx}
\usepackage{pifont}
\usepackage{tikz}
\usepackage{xspace}
\usetikzlibrary{positioning,arrows.meta,calc,fit,backgrounds,shapes.geometric}

\newcommand{\R}{\mathbb{R}}

\newcommand{\sca}{\textsc{SCA}}
\newcommand{\ourmethod}{{\fontfamily{lmtt}\selectfont \textbf{SafeEvo}}\xspace}

\newtcolorbox{sidebox}[1][]{colback=gray!5,colframe=gray!30,boxrule=0.4pt,arc=1mm,left=4pt,right=4pt,top=3pt,bottom=3pt,#1}

\title{SafeEvo: Deciphering the Safety Alignment Mechanism and Evolution in Language Models}

\author{
Miao Yu$^{1, *}$,\; 
Hao Huang$^{2, *}$,\; 
Lu Yuan$^{3}$,\; 
Yunpeng Li$^{2}$,\; 
Kun Wang$^{4}$,\;
Zuming Jiang$^{1, \dagger}$\;\\
$^{1}$The University of Hong Kong (HKU)\quad \\
$^{2}$Chinese Academy of Sciences (CAS)\quad \\
$^{3}$Information Engineering University (IEU)\quad  \\
$^{4}$Nanyang Technological University (NTU)\quad \\
}

\begin{document}
\maketitle

\begingroup
\renewcommand{\thefootnote}{\ensuremath{*}}
\footnotetext{Equal Contribution}
\endgroup

\begingroup
\renewcommand{\thefootnote}{\ensuremath{\dagger}}
\footnotetext{Corresponding Author}
\endgroup

\input{sections/0_abstract}
\input{sections/1_introduction}
\input{sections/2_related_work}

\input{sections/3_method}
\input{sections/4_experiments}
\input{sections/6_conclusion}
\input{sections/7_ai_use_statement}

\bibliographystyle{iclr2027_conference}
\bibliography{references}

\appendix
\input{sections/A_appendix}

\end{document}

%% file: sections/0_abstract.tex
\begin{abstract}
Safety interpretability advances the study of Large Language Model (LLM) alignment from behavioral constraints driven by data or algorithms towards a deeper understanding of internal mechanisms. However, existing works have focused primarily on safety-related representations, attention heads, or neurons after alignment, while largely overlooking the safety mechanisms in pretrained-only models and their evolution across alignment checkpoints. To address this, we propose \ourmethod, an interpretability framework from the circuit (sparse subgraphs of an LLM) perspective. \ourmethod first applies an optimization-based extraction algorithm to identify weak refusal circuits in pretrained base LLMs that can independently express refusal behavior. Causally ablating these circuits completely eliminates the base model’s refusal of harmful inputs. \ourmethod then traces the evolution of refusal circuits across successive alignment checkpoints and finds that their structures change progressively, suggesting that the alignment tax may result from refusal-circuit updates affecting utility-related parameters. To validate this, \ourmethod introduces Safety Circuit Alignment (SCA), which confines safety updates to the refusal circuits. Experiments across three LLMs and two alignment algorithms show that, on average, SCA outperforms vanilla alignment in three aspects: \textbf{(1) stronger alignment}, lowering harmfulness score by 63.21\%; \textbf{(2) less over-refusal}, yielding a 58.44\% decrease in refusal rates for benign queries; and \textbf{(3) better utility}, retaining 99.58\% of the original model capabilities. Our code is available at: \url{https://github.com/a727990982/SafeEvo}.
\end{abstract}

%% file: sections/1_introduction.tex
\section{Introduction}
\label{sec:intro}

As large language models (LLMs) evolve from conversational models~\citep{team2026kimi, xu2026deepseek} into complex agentic systems~\citep{ferrag2026llm, lee2026survey}, their capabilities and applications expand substantially. This transition amplifies the risks of harmful behavior, which can propagate through tool use to cause direct real-world harm~\citep{luo2026agentauditor, yu2025survey, wang2025comprehensive}. Reliable safety alignment is therefore essential. At its core, alignment seeks to steer model behavior toward safety requirements while preserving useful capabilities~\citep{ji2026survey, li2026matters, zhou2025survey}. Existing methods pursue this goal via supervised demonstrations, preference feedback, or reward design~\citep{instructgpt,pku_saferlhf,kim2026reasoning, cao2026reasoned}. Nevertheless, the research in this line focuses on surface-level and behavioral patterns, with limited investigation into the underlying safety mechanisms, despite the corresponding insights for better alignment algorithms and more trustworthy models~\citep{bereska2024mechanistic}.

In fact, another line of works on safety interpretability~\citep{lee2025interpretation} have attempted to explain the internal mechanisms of alignment~\citep{elhage2021mathematical}. By designing heuristic metrics and search algorithms, they identify attention heads~\citep{zheng2026sahs, zhou2025role, zheng2025spot}, neurons~\citep{zhao2025understanding, yi2025nlsr}, layers~\citep{li2025safetylayers}, and representation~\citep{jiao2026llm, luo2026detecting, zhou2024alignment} (e.g., refusal direction but we establish the more interpretable and fundamental refusal circuits) associated with safety concepts. However, these interpretability studies and the related alignment methods \citep{yang2026asft,zhao2025sntune,li2025salora} focus on the already safety-aligned models only at a fixed endpoint. They lack both attention and methodologies to delve into the safety mechanisms in pretrained base LLMs and their evolution dynamics during safety post-training, which may contribute to alignment improvement.


\begin{figure}[t]
\centering
\includegraphics[width=\linewidth]{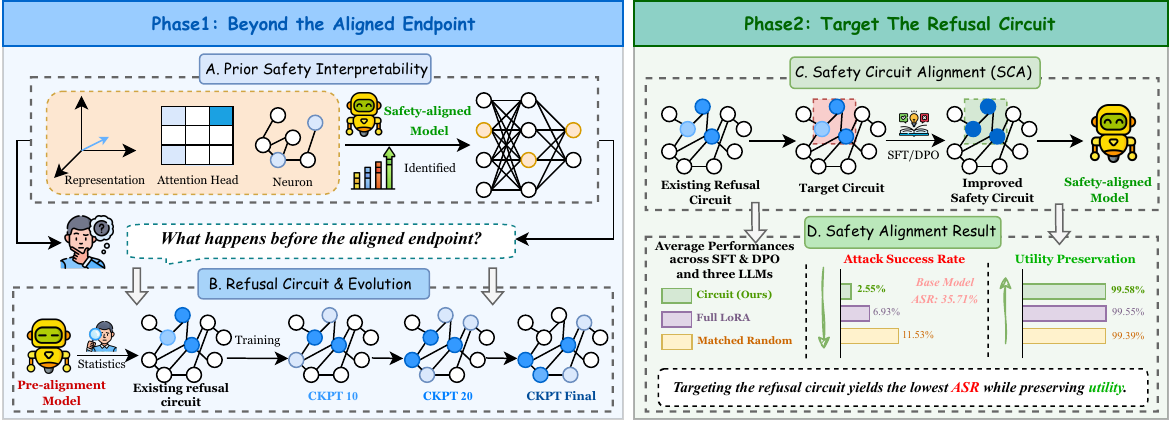}
\vspace{-1.5em}
\caption{\textbf{From circuit discovery to circuit-targeted alignment.}
We move beyond endpoint interpretability by tracing refusal circuits from pretrained-only models and their subsequent evolution (\textit{Phase 1}). We then develop alignment methods that directly target the identified circuits for better safety-utility trade-off (\textit{Phase 2}).}
\label{fig:overview}
\vspace{-1.5em}
\end{figure}

To this end, we introduce \ourmethod, an interpretability framework that attributes LLM alignment to sparse circuits capable of independently eliciting refusal, identifies these refusal circuits in pretrained-only LLMs, and traces their evolution throughout alignment. \ourmethod first casts circuit extraction as an optimization problem rather than relying on heuristic attribution~\citep{behrouzi2026nest, zhou2025role}. Specifically, it learns differentiable masks for each LLM units that automatically isolate a sparse circuit with full safety ability. Applying this technique, we find that pretrained-only \texttt{Llama-3-8B}, \texttt{Qwen-2.5-7B}, and \texttt{Mistral-7B} models are not entirely devoid of safety-related components. Across these models, ablating the identified refusal circuit increases attack success rate (ASR) on HarmBench by 42.76$\%$, 55.98\%, and 45.91\%, respectively. \ourmethod therefore advances and empirically validates the hypothesis that refusal circuits are already present in base LLMs, albeit in a \textit{weak} (reflected by ASR ranging from 32.7$\%$ $\sim$ 48.43 $\%$) form and effective only for a subset of harmful queries. This finding corroborates and advances prior representation-level evidence~\citep{zhou2024alignment} that base LLMs already encode harmfulness and harmlessness from circuit perspective.


We further use \ourmethod to track refusal circuits across alignment checkpoints and quantify their evolution. Warm-starting each extraction from the circuit identified at the preceding checkpoint reveals a gradual structural drift: the overlap between adjacent circuits decreases from $98\%$ at the onset of alignment to $78\%$ after 20 checkpoints. Moreover, independent extraction runs recover multiple circuits that fully and independently reproduce the refusal behavior but share only $26\%$ overlap. Together, these findings demonstrate that refusal circuits are both non-unique and structurally unstable during alignment. This observation offers a possible mechanistic explanation for the ``alignment tax''—the degradation of utility after alignment: \textit{the underconstrained updates of refusal circuits may spill over into utility-relevant parameters, thereby compromising general performance.}


Inspired by these insights and to empirically support them, we introduce \textbf{Safety Circuit Alignment} (SCA), which confines safety-related updates to \textit{only} one identified refusal circuit and \textit{only} optimizes its associated parameters. This design mitigates the observed non-uniqueness and instability of refusal circuit evolution and can be readily combined with existing alignment algorithms. Across three LLMs on HarmBench~\citep{harmbench} and AdvBench~\citep{advbench}, comparing SCA with vanilla alignment via supervised fine-tuning (SFT)~\citep{zhao2025improving} and direct preference optimization (DPO)~\citep{hu2026alignment} reduces average ASR by 2.29$\%$ points ($43.81\%\downarrow$) and 6.47$\%$ points ($74.94\%\downarrow$), respectively, while retaining more utility on GSM8k, MMLU, and HellaSwag. Besides, SCA outperforms a safety-neuron-based baseline and matched-sparsity ablation on random circuits, validating \ourmethod's mechanistic insights as useful foundations for more effective alignment.

In summary, our main contributions can be listed as follows:
\begin{itemize}[leftmargin=*, nosep, itemsep=4pt]
    \item \textbf{Cross-Stage Framework.}
    We propose \ourmethod, an interpretability framework for attributing LLM alignment to refusal circuits and tracking their evolution throughout safety post-training.

    \item \textbf{Novel Insights.}
    We provide causal and experimental evidence that weak refusal circuits already exist in pretrained-only LLMs and reveal their non-uniqueness and instability in vanilla alignment.

    \item \textbf{Circuit-Guided Alignment.}
    We introduce \sca{}, which localizes safety updates to pre-alignment refusal circuits, achieving a favorable balance among safety, utility preservation, and over-refusal.
\end{itemize}

%% file: sections/2_related_work.tex
\section{Related Work}
\label{sec:related}

\textbf{LLM Alignment.} To prevent LLMs from enabling dangerous behavior~\citep{liu2023trustworthy, luo2026survey, zhang2024agent}, prevailing alignment works steer model responses to harmful queries toward refusal, either via SFT~\citep{yang2026asft, fan2026safegen} on synthetic refusal data or via DPO on preference pairs~\citep{ji2026safe, deng2026less}. Optimizing for refusal alone, however, tends to induce over-refusal and erode general capability~\citep{jalan2026survey, niu2026mitigating}. A subsequent line of work mitigates this tension by synthesizing data that is jointly helpful and harmless~\citep{wang2024helpsteer, huang2026explaining}, and by shaping rewards with explicit utility constraints~\citep{ji2026survey}. Orthogonally, mechanistic-interpretability approaches localize neurons causally implicated in the safety boundary~\citep{wang2026safeneuron,zhao2025sntune, nest}. \ourmethod departs from this last line in the granularity and sufficiency of the located structure: we identify a subgraph within LLMs (circuit) that independently realizes safety-refusal behavior — a sufficiency property individual causal neurons do not exhibit. As a more intrinsic locus of the model's safety capability, this circuit reasonably affords a better alignment–utility trade-off under targeted tuning.



\textbf{Safety Interpretability.} Applying perspectives and techniques from mechanistic interpretability~\citep{naseem2026mechanistic} to safety, studies in this direction asks how safety behaviors are computed inside the LLM and why they fail~\citep{lad2024mechanistic, lee2025interpretation}. Prior efforts differ mainly in the granularity at which they localize safety. Representation-level analyses decode intermediate activations and find that refusal versus jailbreak tracks the polarity encoded in hidden states~\citep{he2025interpretable, jiao2026llm}. Component-level studies use activation patching and causal attribution to show that a sparse set of attention heads carries refusal~\citep{zheng2026sahs, zhou2025role}, and that ablating them suffices to remove it. Neuron-level work identifies safety neurons and repurposes them for jailbreak analysis~\citep{zhao2026unraveling} and alignment strengthening~\citep{zhao2025sntune}. These efforts, however, largely examine post-alignment LLMs. \ourmethod instead adopts a circuit-level view~\citep{safeseek} and traces the evolution of safety-related circuits across alignment checkpoints from pretrained-only base LLMs, yielding deeper understanding and insights for designing better alignment algorithms.


%% file: sections/3_method.tex
\vspace{-0.6em}
\section{Refusal Circuit Discovery and Evolution}
\label{sec:method}
\vspace{-0.4em}



Our investigation begins with a simple yet subtle phenomenon: pretrained-only base LLMs refuse a small fraction of harmful queries, as also shown in prior work~\citep{kissane2024baserefuse}. Building on this observation, we hypothesize that pre-alignment LLMs may contain certain separable components responsible for refusal, or more specifically, a sparse and functionally complete refusal circuit.

\begin{figure}[!ht]
\centering
\includegraphics[width=\linewidth]{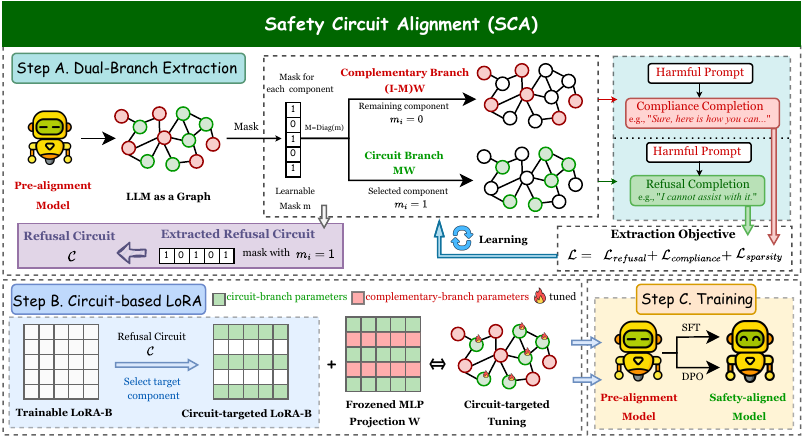}
\vspace{-1.5em}
\caption{\textbf{Overview of our Safety Circuit Alignment algorithm.} It first extracts a refusal circuit before LLM alignment (\textit{Step A}) and only fine-tunes the identified parameters in that circuit for precise alignment (\textit{Step B}).}
\vspace{-1em}
\label{fig:method}
\end{figure}

We first formulate how to identify safety-related refusal
circuits (Section~\ref{sec:circuit-extraction}). Then, we test the causal contribution and ability of these circuits
via ablation and isolation, respectively
(Section~\ref{sec:method-ablate}). Finally, we track refusal circuits during
early safety alignment, examining changes across checkpoints and variation across
independent extractions (Section~\ref{sec:method-drift}).
Figure~\ref{fig:method} demonstrates the overview of \ourmethod and its 
application for better alignment via localized refusal circuit tuning.

\subsection{Refusal Circuit Extraction}
\label{sec:circuit-extraction}
We refer to a sparse subset of model components, together with the computational subgraph they induce, that can express refusal behavior in isolation as a \emph{refusal circuit}. Extracting such a circuit amounts to a selection over a combinatorially large space. A common approach in previous works guides this using heuristic component-level scores based on activations, attributions, or weights~\citep{zhao2025sntune,chen2024safetyneurons,nlsr}, but they only capture circuit \textit{for \textbf{only one} token generation.} We instead cast refusal circuit discovery as an optimization problem: each candidate component is associated with a learnable binary selection variable (a \emph{mask}), and with the LLM weights frozen, the masks are optimized jointly such that the selected express refusal on their own, their complement generates harmful-compliance completions, and the selection remains sparse.

\textbf{Circuit Components.}
We take \emph{MLP neurons} as the unit of refusal circuits for selection, since Multi-layer Perceptron (MLP) modules in LLMs are widely regarded as the primary locus of stored knowledge and behaviors. The gated MLP (common in today's LLMs) at layer $\ell$ can be defined as:
\begin{equation}
\operatorname{MLP}_\ell(h_\ell)=W_{\ell,\mathrm{down}}
\left[\phi(W_{\ell,\mathrm{gate}}h_\ell)\odot
(W_{\ell,\mathrm{up}}h_\ell)\right],
\label{eq:mlp-forward}
\end{equation}
where $h_\ell\in\R^{d_{\mathrm{model}}}$ is the MLP input, $\phi$ is the gating function, and $\odot$ is element-wise multiplication. $W_{\ell,\mathrm{gate}},W_{\ell,\mathrm{up}}\in\R^{d_{\mathrm{ff}}\times d_{\mathrm{model}}}$ map to the intermediate MLP space, while $W_{\ell,\mathrm{down}}\in\R^{d_{\mathrm{model}}\times d_{\mathrm{ff}}}$ maps back to the model space. Each projection is a linear map whose $i$-th output coordinate is computed exclusively by the $i$-th row of its weight matrix. We regard each such row--output pair as one MLP neuron. For each projection $p\in\mathcal P=\{\mathrm{gate},\mathrm{up},\mathrm{down}\}$, we denote its frozen weight, binary neuron mask, and diagonal mask by the following, respectively:
\begin{equation}
W_{\ell,p}\in\R^{d_{\mathrm{out},p}\times d_{\mathrm{in},p}},\qquad
m_{\ell,p}\in\{0,1\}^{d_{\mathrm{out},p}},\qquad
M_{\ell,p}=\operatorname{Diag}(m_{\ell,p}),
\label{eq:mask-definition}
\end{equation}
and collect all candidates into the universal neuron set $\mathcal U=\{(\ell,p,i)\mid \ell\in[L],\,p\in\mathcal P,\,i\in[d_{\mathrm{out},p}]\}$. Since the projections are row-separable, any neuron can be switched off by zeroing its row without altering other neurons, making $\mathcal U$ a natural basis for defining refusal circuits.

\textbf{Refusal Circuit.}
Let $\mathcal M=\{m_{\ell,p}\}_{\ell,p}$ collect all neuron masks and $\overline{\mathcal M}=\{\mathbf 1-m_{\ell,p}\}_{\ell,p}$ denote its complement. For input $h_{\ell,p}$, the original, circuit, and complementary outputs of each projection are:
\begin{equation}
z_{\ell,p}=W_{\ell,p}h_{\ell,p},\qquad
\underbrace{\textcolor{scacircuit}{z^{C}_{\ell,p}=M_{\ell,p}W_{\ell,p}h_{\ell,p}}}
_{\textcolor{scacircuit}{\scriptstyle\text{circuit for refusal}}},\qquad
\underbrace{\textcolor{scacomplement}{z^{\bar C}_{\ell,p}=(I-M_{\ell,p})W_{\ell,p}h_{\ell,p}}}
_{\textcolor{scacomplement}{\scriptstyle\text{complement for compliance}}}.
\label{eq:circuit-forward}
\end{equation}
Since $M_{\ell,p}+(I-M_{\ell,p})=I$, we have $z_{\ell,p}=z^{C}_{\ell,p}+z^{\bar C}_{\ell,p}$, i.e., the masks induce an exact partition of $\mathcal U$ into a circuit part (sparse refusal circuit) and a complement part (the remaining dense graph):
\begin{equation}
C=\{(\ell,p,i)\mid m_{\ell,p,i}=1\},\qquad
\bar C=\mathcal U\setminus C,\qquad
|C|=\sum_{\ell,p}\lVert m_{\ell,p}\rVert_0,
\label{eq:circuit-set}
\end{equation}
where $|C|$ counts the selected neurons. The \emph{circuit model} $p_{\theta,\mathcal M}$ replaces every projection output in Eq.~\ref{eq:mlp-forward} by $z^{C}_{\ell,p}$, zero-ablating all neurons in $\bar C$; symmetrically, the \emph{complementary model} $p_{\theta,\overline{\mathcal M}}$ uses $z^{\bar C}_{\ell,p}$ and zero-ablates $C$. Masking a gate or up neuron removes its intermediate coordinate from the input of $W_{\ell,\mathrm{down}}$, and masking a down neuron removes the MLP's contribution to one residual-stream coordinate, so $C$ specifies a sparse computational subgraph rather than a bag of isolated neurons.

\textbf{Differentiable Mask Optimization.}
The binary masks in Eq.~\ref{eq:mask-definition} are not differentiable. We therefore parameterize each mask by a latent vector $q_{\ell,p}\in\R^{d_{\mathrm{out},p}}$ with soft mask $\widetilde m_{\ell,p}=\sigma(q_{\ell,p})\in(0,1)^{d_{\mathrm{out},p}}$. A purely continuous relaxation, however, lets the circuit rely on fractionally scaled neurons that vanish once thresholded, causing a train--test inconsistency. To keep the forward pass strictly binary and differentiable, we binarize the soft masks with threshold $\eta$ via a straight-through estimator (STE):
\begin{equation}
m_{\ell,p}=\mathbb 1\!\left[\widetilde m_{\ell,p}>\eta\right],\qquad
M_{\ell,p}=\operatorname{Diag}\!\left(\widetilde m_{\ell,p}
+\operatorname{sg}(m_{\ell,p}-\widetilde m_{\ell,p})\right),
\label{eq:ste-mask}
\end{equation}
where $\operatorname{sg}(\cdot)$ is the stop-gradient operator. In the forward pass $M_{\ell,p}$ equals the binary $\operatorname{Diag}(m_{\ell,p})$, so Eq.~\ref{eq:circuit-forward} always evaluates a discrete partition; in the backward pass, $\nabla\operatorname{sg}(\cdot)\equiv 0$ routes the gradient of $M_{\ell,p}$ to $\widetilde m_{\ell,p}$ and hence to $q_{\ell,p}$. We initialize $q_{\ell,p}$ with large positive values so that $M_{\ell,p}=I$, i.e., extraction starts from the intact LLM and progressively prunes neurons out of the circuit.

\paragraph{Dual-branch Extraction Objective.}
We learn the masks with two completion-supervision branches: the refusal branch trains the circuit on refusal completions $\mathcal D_{\mathrm{ref}}$, whereas the compliance branch trains the complementary model on harmful-compliance completions $\mathcal D_{\mathrm{cmp}}$. The total loss is:
\begin{align}
\mathcal L_{\mathrm{ext}}(\mathcal M)
={}&\underbrace{\textcolor{scacircuit}{
\alpha\,\mathcal L_{\mathrm{SFT}}(\mathcal M;\mathcal D_{\mathrm{ref}})}}
_{\textcolor{scacircuit}{\scriptstyle\text{circuit / refusal}}}
+\underbrace{\textcolor{scacomplement}{
\beta\,\mathcal L_{\mathrm{SFT}}(\overline{\mathcal M};\mathcal D_{\mathrm{cmp}})}}
_{\textcolor{scacomplement}{\scriptstyle\text{complement / compliance}}}
+\lambda_{\mathrm{mlp}}\sum_{\ell,p}
\frac{\lVert\widetilde m_{\ell,p}\rVert_1}{d_{\mathrm{out},p}},
\label{eq:extraction-loss}
\end{align}
where $\mathcal L_{\mathrm{SFT}}(\mathcal M;\mathcal D)$ is the token-level negative log-likelihood of completions $y$ given prompts $x$ in $\mathcal D$ under the masked model $p_{\theta,\mathcal M}$, and only $\{q_{\ell,p}\}$ are optimized while $\theta$ remains frozen. The two branches assign complementary roles to the partition: the refusal term requires $C$ to be sufficient for refusal, while the compliance term requires $\bar C$ to retain harmful compliance, ruling out the trivial solution $M_{\ell,p}=I$ and implying that removing $C$ suppresses refusal. The normalized $L_1$ penalty on the soft masks prunes uninformative neurons, yielding a compact circuit; $\alpha$, $\beta$, and $\lambda_{\mathrm{mlp}}$ balance the three terms. Upon convergence, the refusal circuit is read out from the binary masks via Eq.~\ref{eq:circuit-set}.

\begin{table}[!t]
\centering
\small
\caption{\textbf{Refusal rate of the base model, the refusal circuit and its complement branch for harmful queries in the BeaverTails dataset.} Marker \textcolor{scacircuit}{\scriptsize$\uparrow$} and \textcolor{scacomplement}{\scriptsize$\downarrow$} show the increase and decrease compared with the base models.}
\vspace{-1em}
\label{tab:circuit-sufficiency}
\begin{adjustbox}{width=1\linewidth}
\begin{tabular}{lcccc}
\toprule
\rowcolor{gray!80!blue!40}
\textbf{Models / Refusal Rate (\%)} & \textbf{Base Model} &
\textbf{Refusal Circuit ($C_1$)} & \textbf{Density ($C_1$)} &
\textbf{Complement Branch} \\
\midrule
Llama-3-8B & 48.04 &
\textbf{100.00}\,\textcolor{scacircuit}{\scriptsize$\uparrow 51.96$} &
0.74\% &
\textbf{0.00}\,\textcolor{scacomplement}{\scriptsize$\downarrow 48.04$} \\
Qwen-2.5-7B & 35.15 &
\textbf{100.00}\,\textcolor{scacircuit}{\scriptsize$\uparrow 64.85$} &
0.80\% &
\textbf{0.00}\,\textcolor{scacomplement}{\scriptsize$\downarrow 35.15$} \\
Mistral-7B-v0.1 & 15.63 &
\textbf{100.00}\,\textcolor{scacircuit}{\scriptsize$\uparrow 84.37$} &
0.75\% &
\textbf{0.00}\,\textcolor{scacomplement}{\scriptsize$\downarrow 15.63$} \\
\bottomrule
\end{tabular}
\end{adjustbox}
\end{table}


\begin{figure}[!t]
\centering
\includegraphics[width=\linewidth]{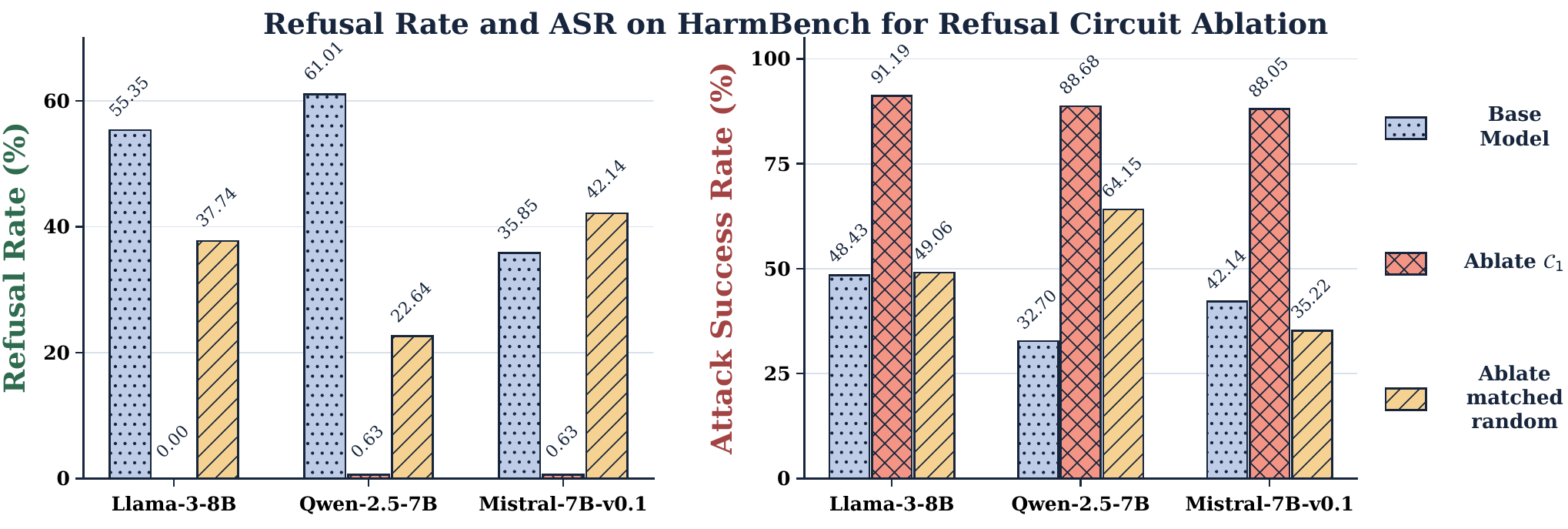}
\vspace{-2em}
\caption{\textbf{Refusal rate and ASR on HarmBench for base LLMs and the same models with the identified refusal circuits $C_1$ ablated, compared to the density-matched random ablation models.}}
\label{fig:ablation}
\vspace{-1em}
\end{figure}

\subsection{Refusal Circuits in Pretrained-only LLMs}
\label{sec:method-ablate}
\label{sec:exp-exists}

We apply the techniques in Section~\ref{sec:circuit-extraction} and LLM-LAT~\citep{llmlat} as the dataset to extract a candidate refusal circuit $C_1$ from \texttt{Llama-3-8B}, \texttt{Qwen-2.5-7B}, and
\texttt{Mistral-7B-v0.1}, all containing only < 1\% of the model parameters. Then we investigate the properties of the base LLMs, the circuits, and their complement branches via two behavioral metrics: ASR, which measures positive compliance for harmful queries, and refusal rate, which measures explicit refusal behaviors.

\textbf{Takeaway 1: Our extracted candidates are sparse, functionally complete refusal circuits that can express refusal in isolation.} We evaluate the refusal rates on BeaverTails~\citep{pku_saferlhf}, a harmful-query dataset that is \textit{out-of-distribution} (OOD) relative to the data for circuit search. As shown in Table~\ref{tab:circuit-sufficiency}, pretrained-only base LLMs occasionally refuse harmful queries (with refusal rates of $15\% \sim 48\%$), whereas every extracted circuit achieves a 100\% OOD refusal rate while containing only < $1\%$ parameters. These results indicate that our extraction method successfully identifies a refusal circuit that generalizes to unseen harmful queries. Moreover, ablating the circuit eliminates refusal in the complete branch with all refusal rates drops to $0\%$. \textit{Together, these findings suggest that a sparse circuit plays the core and fundamental role in LLM refusal of harmful requests.}


\textbf{Takeaway 2: Ablating refusal circuits substantially increases model harmfulness.} To further test the causal relationship between refusal circuits and safety alignment, we zero ablate the outputs of components in each refusal circuit and compare the results with a random zero-ablation baseline at the same density. As shown in Figure~\ref{fig:ablation}, circuit ablation severely impairs refusal, yielding refusal rates of $0.00\%\sim0.63\%$. Meanwhile, ASR rises from 48.43\%, 32.70\%, and 42.14\% to 91.19\%, 88.68\%, and 88.05\%, respectively (relative increases of 88.3\%, 171.2\%, and 109.0\%). The increases under random ablation are much smaller. \textit{These causal ablation results show that safety alignment is causally linked to refusal circuits, suggesting that LLMs’ safety capabilities may originate from them.}

\begin{figure}[!t]
\centering
\includegraphics[height=4.6cm]{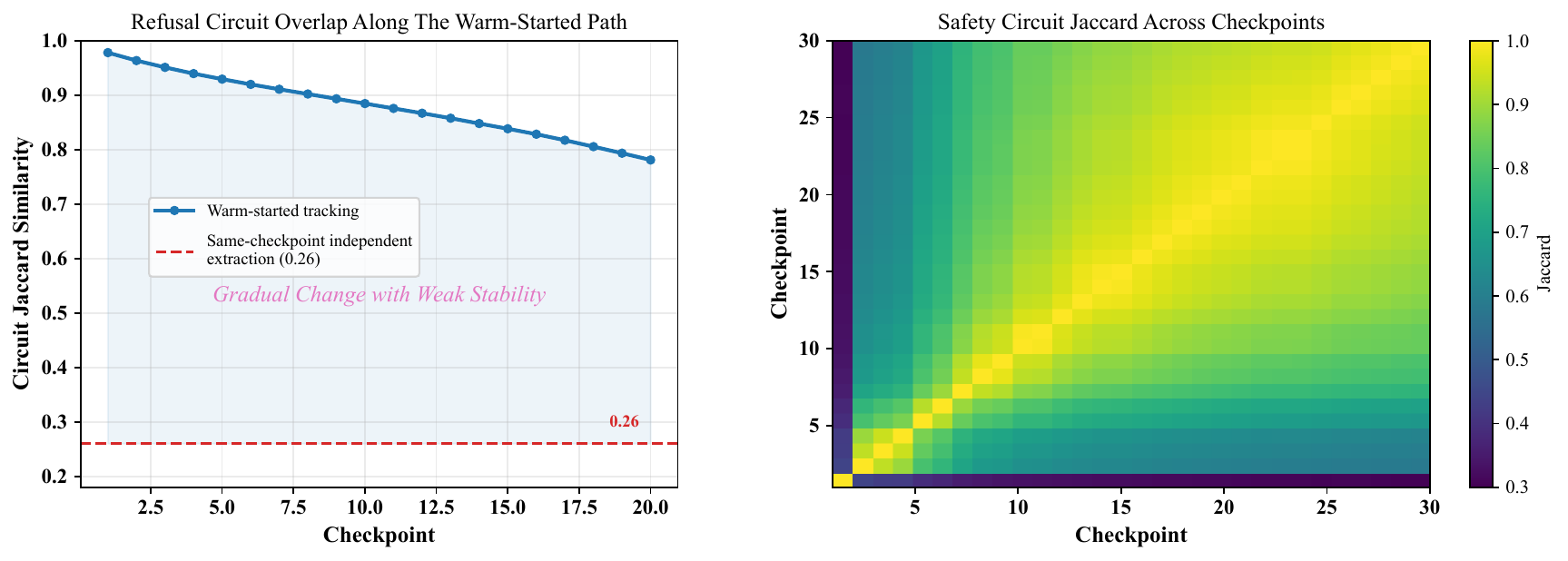}
\vspace{-1em}
\caption{\textbf{Evolution of refusal circuits across alignment, traced and quantified by Jaccard similarity.}}
\label{fig:drift}
\vspace{-1.5em}
\end{figure}

\subsection{Refusal Circuit Evolution During Safety Alignment}
\label{sec:method-drift}
\label{sec:exp-drift}

Section~\ref{sec:method-ablate} provides experimental evidence for functional refusal circuits in base LLMs, but does not investigate their evolution (whether the same circuits still support refusal as alignment proceeds). We therefore track the composition of extracted refusal circuits across safety-alignment checkpoints.

\textbf{Tracking protocol.} We apply SFT-based safety alignment to the \texttt{Llama-3-8B} base model and track the evolution of its refusal circuit over the first several aligned checkpoints. To capture incremental changes, we warm-start each extraction from the circuit found at the preceding checkpoint. Then we measure refusal circuit continuity using the Jaccard overlap between consecutive checkpoints. Besides, we also perform multiple independent extractions at a single checkpoint to further quantify run-to-run variability, providing a reference for assessing cross-checkpoint similarity.

\textbf{Takeaway 3: Refusal circuits are not unique within a given LLM or checkpoint.} As shown in the left panel of Figure~\ref{fig:drift}, independently extracted refusal circuits at the same checkpoint exhibit only 0.26 average component overlap, far below the $0.78\sim0.98$ similarity for warm-started extractions. \textit{This indicates that refusal functionality is implemented redundantly in LLMs, with multiple valid refusal circuits sharing only certain components (maybe relate to basic language capabilities).}


\textbf{Takeaway 4: Refusal circuits undergo gradual structural drift during alignment.} As shown by the warm-started extractions in Figure~\ref{fig:drift}, component similarity between consecutive checkpoints declines smoothly from $1.00 \to 0.78$, indicating that the circuits progressively acquire and discard components rather than remaining fixed. Cross-checkpoint comparisons (the right similarity heatmap of Figure~\ref{fig:drift}) further reveal a larger shift between the base model’s (pre-alignment) circuit and those after several alignment steps (similarity $\sim 0.60$). \textit{These results consistently support the gradual evolution and weak stability of refusal circuits during LLM safety post-training.}



Together, these two findings establish a simple picture of refusal circuits: refusal originates in and can be attributed to different circuits whose composition evolves gradually during alignment. This view may offers a potential account of the ``alignment tax'' incurred during safety post-training: \textit{safety alignment alters the refusal circuits that underlie LLMs’ refusal capabilities, and unconstrained parameter updates may inadvertently affect parameters associated with utility, including general capabilities.} Motivated by this hypothesis and seeking to test it empirically to some extent, we propose restricting safety-alignment updates entirely to the refusal circuit in the base LLMs, which is already present and does not interfere with the knowledge and capabilities acquired during pretraining.


%% file: sections/4_experiments.tex
\section{Safety Circuit Alignment}
\label{sec:exp}

To test the above hypothesis and develop a more effective alignment algorithm, we introduce Safety Circuit Alignment (\sca{}), which restricts safety updates to the pre-alignment refusal circuits in base LLMs, inspired by our interpretability insights in Section \ref{sec:method}. We first formulate the core principle of SCA in Section \ref{sec:method-sca} and introduce the experiment setups for subsequent safety alignment in Section \ref{sec:method-controls}. Then we present and analyze the experiment results of our methods with other baselines' in Section \ref{sec:exp-align}. The overview of our SCA technique is illustrated in Figure \ref{fig:method}.


\subsection{Alignment Algorithm (SCA)}
\label{sec:method-sca}

\sca{} confines all safety-alignment updates to the refusal circuit $C$ extracted from the pre-alignment LLM. Concretely, we parameterize the update of each MLP projection with LoRA and reuse the binary neuron mask $M_{\ell,p}=\operatorname{Diag}(m_{\ell,p})$ obtained by Eq.~\ref{eq:extraction-loss} to restrict the rank-$r$ update to the circuit:
\begin{equation}
W'_{\ell,p}=W_{\ell,p}+M_{\ell,p}B_{\ell,p}A_{\ell,p},
\label{eq:sca}
\end{equation}
where $A_{\ell,p}\in\R^{r\times d_{\mathrm{in},p}}$ and $B_{\ell,p}\in\R^{d_{\mathrm{out},p}\times r}$. Since $M_{\ell,p}$ is diagonal and binary, $M_{\ell,p}B_{\ell,p}A_{\ell,p}$ has non-zero rows exactly at the neurons in $C$ (Eq.~\ref{eq:circuit-set}), so $W'_{\ell,p}$ differs from $W_{\ell,p}$ only on the rows of circuit neurons, while the weights of $\bar C$ and all attention modules remain identical to the base model. During alignment, $M_{\ell,p}$ stays fixed and $W_{\ell,p}$ is frozen; only $A_{\ell,p}$ and $B_{\ell,p}$ are optimized.

\sca{} changes the update support rather than the alignment loss. Let $o\in\{\mathrm{SFT},\mathrm{DPO}\}$ index the objective and $\theta_M^{(k)}$ denote the policy parameterized by Eq.~\ref{eq:sca} at step $k$. \sca{}'s update is formulated as:
\begin{equation}
B_{\ell,p}^{(k+1,o)}
=B_{\ell,p}^{(k)}-
\gamma\,M_{\ell,p}\nabla_{B_{\ell,p}}\mathcal L_o(\theta_M^{(k)}),
\label{eq:sca-update}
\end{equation}
where $\gamma$ is the learning rate; the input factor $A_{\ell,p}$ follows the ordinary optimizer update. $M_{\ell,p}$ gates the rows of the gradient so that neurons outside $C$ never receive an update, mirroring how Eq.~\ref{eq:circuit-forward} gates their outputs during extraction. We instantiate $\mathcal L_o$ with two prevailing alignment objectives. For SFT, we adopt the $\mathcal L_{\mathrm{SFT}}$ used in Section~\ref{sec:circuit-extraction}, now applied to $\pi_{\theta_M}$ on refusal completions. For DPO, given a prompt $x$ with a safe completion $y_w$ and a harmful completion $y_l$, we optimize:
\begin{equation}
\mathcal L_{\mathrm{DPO}}(\theta_M)
=-\mathbb E_{(x,y_w,y_l)}\log\sigma\!\left(\tau\!\left[
\log\frac{\pi_{\theta_M}(y_w\mid x)}{\pi_{\mathrm{ref}}(y_w\mid x)}-
\log\frac{\pi_{\theta_M}(y_l\mid x)}{\pi_{\mathrm{ref}}(y_l\mid x)}
\right]\!\right),
\label{eq:sca-dpo}
\end{equation}
where $\pi_{\mathrm{ref}}$ is the frozen pre-alignment model and $\tau$ controls the deviation from it. Both losses are computed through the full forward pass of the LLM, but its gradient only alters the refusal circuits.

\subsection{Experimental Settings}
\label{sec:method-controls}

\textbf{Models.} We select three LLMs: \texttt{Llama-3-8B}~\citep{llama3}, \texttt{Qwen-2.5-7B}~\citep{qwen25}, and \texttt{Mistral-7B-Instruct-v0.1}~\citep{mistral7b}. Notably, we use the Mistral's instruction-tuned version because its base model does not reliably follow the instruction format.

\textbf{Alignment Objectives \& Datasets.} We utilize two prevailing and most important losses in safety aligment: SFT and DPO. Specifially, SFT uses the refusal completions from LLM-LAT~\citep{llmlat}, while DPO adopts safety preference pairs in PKU-SafeRLHF~\citep{pku_saferlhf}.
 
\textbf{Evaluation Metrics \& Datasets.} For the safety evaluation, we calculate attack success rate (ASR) via the Llama-Guard-3-8B~\citep{llamaguard3} model on HarmBench~\citep{harmbench} and AdvBench~\citep{advbench}. LLMs' general utility after safety alignment is measured by the accuracies on GSM8K~\citep{gsm8k}, MMLU~\citep{mmlu}, and HellaSwag~\citep{hellaswag}, while over-refusal evaluation adopts the benign input prompts in XSTest~\citep{xstest} and BeaverTails~\citep{ji2023beavertails}. Notably, all evaluation datasets are held out from SFT and DPO training, preventing direct fitting and enabling a rigorous assessment of OOD generalization.

\textbf{Baselines.} All alignment methods in our experiments use either the SFT or DPO loss and differ only in how parameters are updated. As the basic baseline, we use vanilla fine-tuning via LoRA, where all model parameters are subject to updating. To isolate the effect of sparse updates in SCA, we compare it with a randomly selected update method at the same sparsity level, which does not specifically target refusal circuits. To demonstrate SCA's deeper interpretability insights, we also include SN-Tune~\citep{zhao2025sntune}, an update method based on the interpretability of safety neurons as an additional baseline. Implementation details are provided in Appendix~\ref{app:sn-baseline}.

\input{sections/tables_v6}

\subsection{Main Safety--Utility Results}
\label{sec:exp-align}

To validate SCA, developed based on our interpretability insights, we evaluate aligned LLMs on safety and utility datasets that are OOD to training, with results in Table~\ref{tab:main} and \ref{tab:ood}. On average, SCA outperforms baselines in alignment efficacy, capability preservation, and avoiding over-refusal.


\textbf{Better Alignment.} As shown in Table~\ref{tab:main}, SCA provides strong safety improvements under both SFT and DPO. Across all six model--objective blocks, its HarmBench and AdvBench attack success rates rank either first or second among full-LoRA, random-support, and SCA. Under DPO, SCA reduces the Llama-3-8B ASR $\to$ \(0.63\%\) on HarmBench and \(0.38\%\) on AdvBench, compared with \(16.98\%\) and \(12.69\%\) for full-LoRA. It also achieves \(7.55\%\) on HarmBench and \(4.23\%\) on AdvBench for Mistral-7B-Instruct-v0.1, substantially below full-LoRA's \(16.35\%\) and \(5.58\%\). Under Qwen-2.5-7B SFT, SCA keeps the two ASRs at \(6.92\%\) and \(0.38\%\), while random-support reaches \(20.75\%\) and \(2.50\%\). These results suggest that SCA effectively maintains robust refusal behavior on held-out harmful queries, supporting our interpretability insights in safety alignment and refusal circuits.

\textbf{Utility Preservation.} At the same time, SCA preserves the models' general capabilities across GSM8K, MMLU, and HellaSwag. Across the six model--objective blocks, the largest utility decrease is only \(1.44\)\% points, while several benchmarks improve over the starting model. For instance, under SFT, Mistral-7B-Instruct-v0.1 changes only from \(33.43\) $\to$ \(32.68\) on GSM8K, from \(53.74\) $\to$ \(53.75\) on MMLU, and from \(50.15\) $\to$ \(50.20\) on HellaSwag. Likewise, Qwen-2.5-7B under SFT improves on GSM8K from \(83.09\) $\to$ \(83.17\), while its MMLU score decreases by only \(0.16\) points. Under DPO, Llama-3-8B even improves from \(65.36\) $\to$ \(65.60\) on MMLU and from \(53.05\) $\to$ \(53.45\) on HellaSwag. This stable utility profile indicates that SCA strengthens safety while largely preserving broad abilities, validating our explanation for alignment tax in Section \ref{sec:method-drift} to some extend.

\textbf{Less Over-refusal.} For benign prompts, SCA also substantially reduces over-refusal on XSTest compared with full-LoRA. On Llama-3-8B under SFT, the over-refusal rate increases only from the base value of \(0.8\%\) to \(1.6\%\), whereas full-LoRA reaches \(18.8\%\). The corresponding rates for Qwen-2.5-7B are \(6.8\%\) for SCA and \(28.4\%\) for full-LoRA under SFT, and \(17.6\%\) for SCA and \(34.8\%\) for full-LoRA under DPO. For Mistral-7B-Instruct-v0.1 under SFT, SCA limits over-refusal to \(5.6\%\), compared with \(7.6\%\) for full-LoRA. These results show that SCA preserves responsiveness to benign requests while avoiding the excessive refusal behavior introduced by dense parameter updates.


\begin{table}[t]
\centering
\caption{\textbf{Over-refusal and ASR trade-off on XSTest datasets.} Marker meanings are the same as above.}
\vspace{-1em}
\label{tab:ood}
\resizebox{\textwidth}{!}{%
\begin{tabular}{lccccccc}
\toprule
\rowcolor{gray!80!blue!40}
\textbf{Method} & \textbf{Objective} &
\multicolumn{2}{c}{\textbf{Llama-3-8B}} &
\multicolumn{2}{c}{\textbf{Qwen-2.5-7B}} &
\multicolumn{2}{c}{\textbf{Mistral-7B-v0.1}} \\
\cmidrule(lr){3-4}\cmidrule(lr){5-6}\cmidrule(l){7-8}
& & \textcolor{red}{ASR ($\downarrow$)} & \textcolor{blue}{Over Refusal ($\downarrow)$} &
\textcolor{red}{ASR ($\downarrow$)} & \textcolor{blue}{Over Refusal ($\downarrow)$} & \textcolor{red}{ASR ($\downarrow$)} & \textcolor{blue}{Over Refusal ($\downarrow)$} \\
\midrule
Base Model & -- & \basepadone{64.5} & \basepadone{0.8} &
\basepadone{61.5} & \basepadone{5.2} & \basepadone{66.0} &
\basepadone{0.8} \\
LoRA & \cellcolor{orange!18}\textcolor{BurntOrange}{\textbf{SFT}} &
\textbf{50.5}\sdec{14.0} & \underline{18.8}\xinc{18.0} & \textbf{30.5}\sdec{31.0} & \underline{28.4}\xinc{23.2} & \textbf{40.0}\sdec{26.0} & \underline{7.6}\xinc{6.8} \\
\textbf{SCA (Ours)} & \cellcolor{orange!18}\textcolor{BurntOrange}{\textbf{SFT}} &
\underline{62.0}\sdec{2.5} & \textbf{1.6}\xinc{0.8} & \underline{47.5}\sdec{14.0} & \textbf{6.8}\xinc{1.6} & \underline{51.0}\sdec{15.0} & \textbf{5.6}\xinc{4.8} \\
\cmidrule(lr){1-8}
LoRA & \cellcolor{cyan!14}\textcolor{RoyalBlue}{\textbf{DPO}} &
\underline{53.0}\sdec{11.5} & \textbf{2.8}\xinc{2.0} & \textbf{1.0}\sdec{60.5} & \underline{34.8}\xinc{29.6} & \underline{26.5}\sdec{39.5} & \textbf{4.8}\xinc{4.0} \\
\textbf{SCA (Ours)} & \cellcolor{cyan!14}\textcolor{RoyalBlue}{\textbf{DPO}} &
\textbf{48.5}\sdec{16.0} & \underline{3.2}\xinc{2.4} & \underline{36.0}\sdec{25.5} & \textbf{17.6}\xinc{12.4} & \textbf{21.5}\sdec{44.5} & \underline{5.6}\xinc{4.8} \\
\bottomrule
\end{tabular}
}
\vspace{-0.5em}
\end{table}

\begin{table}[!t]
\centering
\small
\setlength{\tabcolsep}{0pt}
\caption{\textbf{HarmBench ASR of multiple runs of SCA and the random baseline.} ASR is reported as mean$\pm$std over four training
seeds. The $\Delta$ column denotes the decrease of SCA ASR (relative to the random baseline), with brackets
reporting the paired prompt-level $95\%$ bootstrap confidence interval (CI).}
\vspace{-1em}
\label{tab:robustness}
\begin{tabular}{@{}
>{\hspace{6pt}\raggedright\arraybackslash}p{0.22\textwidth}
>{\centering\arraybackslash}p{0.26\textwidth}
>{\centering\arraybackslash}p{0.26\textwidth}
>{\centering\arraybackslash}p{0.26\textwidth}
@{}}
\toprule
\rowcolor{gray!80!blue!40}
\textbf{Models} &
\textbf{SCA ASR (Ours)$\downarrow$} &
\textbf{Random Baseline ASR$\downarrow$} &
\textbf{$\Delta$ (95\% CI)} \\
\midrule
Llama-3-8B
& $\mathbf{5.48\pm3.27}$
& $20.28\pm6.00$
& $\mathbf{14.80}$ [$11.48,18.55$] \\

Qwen-2.5-7B
& $\mathbf{11.32\pm5.51}$
& $19.81\pm3.43$
& $\mathbf{8.49}$ [$4.72,12.42$] \\

Mistral-7B-v0.1
& $\mathbf{5.35\pm2.44}$
& $12.58\pm7.00$
& $\mathbf{7.23}$ [$3.77,10.85$] \\
\bottomrule
\end{tabular}
\vspace{-1em}
\end{table}

\begin{figure}[!t]
\centering
\includegraphics[width=\linewidth]{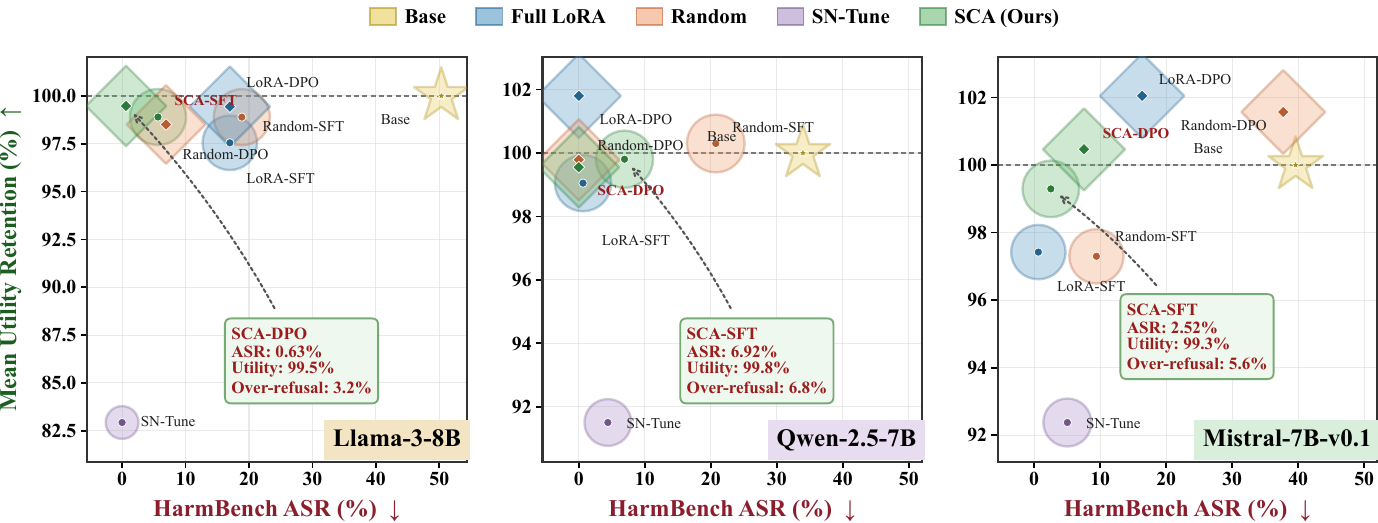}
\vspace{-1.5em}
\caption{\textbf{Safety--utility trade-offs across different LLM families and alignment methods.}}
\label{fig:pareto}
\vspace{-1.5em}
\end{figure}



\subsection{SCA versus Random Baseline}
\label{sec:exp-controls}

We next test whether SCA's advantage in safety alignment persists across training seeds and stems from targeting the extracted refusal circuit instead of sparse updates. Random baseline updates the same amount (as SCA) of but random-selected parameters. Results are displayed in Table~\ref{tab:robustness}.

\textbf{Consistent and Robust Performances.} SCA achieves lower HarmBench ASR than the random baseline for all three models:
$5.48\pm3.27\%$ versus $20.28\pm6.00\%$ for Llama-3-8B,
$11.32\pm5.51\%$ versus $19.81\pm3.43\%$ for Qwen-2.5-7B, and
$5.35\pm2.44\%$ versus $12.58\pm7.00\%$ for Mistral-7B-Instruct-v0.1.
The corresponding ASR reductions are $14.80$, $8.49$, and $7.23$ percentage
points, with all paired prompt-level $95\%$ bootstrap CIs excluding zero.
Because the two conditions use the same update parameterization and sparsity
budget, these results indicate that SCA's stronger safety alignment is not
explained by restricting the update space alone. Instead, the consistent
advantage of SCA supports the relevance of the previously identified refusal
circuit and shows that its circuit-targeted updates provide a more effective
and interpretable support for safety alignment.



%% file: sections/tables_v6.tex

\newcommand{\sdec}[1]{\,{\textcolor{BrickRed}{\tiny$\downarrow#1$}}}
\newcommand{\sinc}[1]{\,{\textcolor{BrickRed}{\tiny$\uparrow#1$}}}
\newcommand{\udec}[1]{\,{\textcolor{ForestGreen}{\tiny$\downarrow#1$}}}
\newcommand{\uinc}[1]{\,{\textcolor{ForestGreen}{\tiny$\uparrow#1$}}}
\newcommand{\usame}{\,{\textcolor{ForestGreen}{\tiny$\rightarrow0.00$}}}
\newcommand{\xinc}[1]{\,{\textcolor{RoyalBlue}{\tiny$\uparrow#1$}}}
\newcommand{\basepadtwo}[1]{#1\phantom{\sdec{00.00}}}
\newcommand{\basepadone}[1]{#1\phantom{\xinc{00.0}}}
\begin{table}[!ht]
\centering\scriptsize
\setlength{\tabcolsep}{3.0pt}
\renewcommand{\arraystretch}{1.04}
\caption{\textbf{Safety, utility, and over-refusal for different baselines under SFT and DPO.}
Colored marker $\downarrow$ and $\uparrow$ indicate changes compared to base models, while those following dataset names denote whether lower or higher values are better. \textbf{Bold} and \underline{underline} mark the best and second-best within each model--objective block.}
\vspace{-1.5em}
\label{tab:main}
\resizebox{\textwidth}{!}{%
\begin{tabular}{lcrrrrrr}
\toprule
\rowcolor{gray!80!blue!40}
\multicolumn{1}{c}{\textbf{Method}} & \multicolumn{1}{c}{\textbf{Objective}} &
\multicolumn{2}{c}{\textbf{Safety}} &
\multicolumn{3}{c}{\textbf{Utility}} &
\multicolumn{1}{c}{\textbf{Over-refusal}} \\
\cmidrule(lr){3-4}\cmidrule(lr){5-7}\cmidrule(l){8-8}
& & \textcolor{BrickRed}{HarmB. ($\downarrow$)} & \textcolor{BrickRed}{AdvB.$ (\downarrow$)} & \textcolor{ForestGreen}{GSM8K ($\uparrow$)} & \textcolor{ForestGreen}{MMLU ($(\uparrow$)} & \textcolor{ForestGreen}{HSwag ($\uparrow$)} & \textcolor{NavyBlue}{XSTest ($\downarrow$)} \\
\midrule
\rowcolor{violet!4}\multicolumn{8}{c}{\textbf{LLAMA-3-8B}} \\
Base Model         & --  & \basepadtwo{50.31} & \basepadtwo{52.88} & \basepadtwo{50.34} & \basepadtwo{65.36} & \basepadtwo{53.05} & \basepadone{0.8} \\
LoRA    & \cellcolor{orange!18}\textcolor{BurntOrange}{\textbf{SFT}} & \underline{16.98}\sdec{33.33} & 12.50\sdec{40.38} & 45.79\udec{4.55} & 63.75\udec{1.61} & \textbf{55.25}\uinc{2.20} & 18.8\xinc{18.0} \\
Random       & \cellcolor{orange!18}\textcolor{BurntOrange}{\textbf{SFT}} & 18.87\sdec{31.44} & \underline{3.27}\sdec{49.61} & \textbf{49.05}\udec{1.29} & \underline{64.49}\udec{0.87} & \underline{53.35}\uinc{0.30} & \underline{2.8}\xinc{2.0} \\
SN-Tune & \cellcolor{orange!18}\textcolor{BurntOrange}{\textbf{SFT}} & 0.00\sdec{50.31} & 0.00\sdec{52.88} & 31.54\udec{18.80} & 57.78\udec{7.58} & 51.85\udec{1.20} & 86.4\xinc{85.6} \\
\textbf{SCA (Ours)} & \cellcolor{orange!18}\textcolor{BurntOrange}{\textbf{SFT}} & \textbf{5.66}\sdec{44.65} & \textbf{1.54}\sdec{51.34} & \underline{48.90}\udec{1.44} & \textbf{64.88}\udec{0.48} & 53.20\uinc{0.15} & \textbf{1.6}\xinc{0.8} \\
\cmidrule(lr){1-8}
LoRA    & \cellcolor{cyan!14}\textcolor{RoyalBlue}{\textbf{DPO}} & 16.98\sdec{33.33} & 12.69\sdec{40.19} & \textbf{49.51}\udec{0.83} & \underline{65.46}\uinc{0.10} & 52.95\udec{0.10} & \textbf{2.8}\xinc{2.0} \\
Random       & \cellcolor{cyan!14}\textcolor{RoyalBlue}{\textbf{DPO}} & \underline{6.92}\sdec{43.39} & \underline{5.00}\sdec{47.88} & 47.99\udec{2.35} & 65.11\udec{0.25} & \underline{53.35}\uinc{0.30} & 3.6\xinc{2.8} \\
\textbf{SCA (Ours)} & \cellcolor{cyan!14}\textcolor{RoyalBlue}{\textbf{DPO}} & \textbf{0.63}\sdec{49.68} & \textbf{0.38}\sdec{52.50} & \underline{48.98}\udec{1.36} & \textbf{65.60}\uinc{0.24} & \textbf{53.45}\uinc{0.40} & \underline{3.2}\xinc{2.4} \\
\midrule
\rowcolor{violet!4}\multicolumn{8}{c}{\textbf{QWEN-2.5-7B}} \\
Base Model         & --  & \basepadtwo{33.96} & \basepadtwo{7.50} & \basepadtwo{83.09} & \basepadtwo{74.21} & \basepadtwo{52.45} & \basepadone{5.2} \\
LoRA    & \cellcolor{orange!18}\textcolor{BurntOrange}{\textbf{SFT}} & \textbf{0.63}\sdec{33.33} & \textbf{0.00}\sdec{7.50} & 80.52\udec{2.57} & 71.91\udec{2.30} & \textbf{54.20}\uinc{1.75} & 28.4\xinc{23.2} \\
Random       & \cellcolor{orange!18}\textcolor{BurntOrange}{\textbf{SFT}} & 20.75\sdec{13.21} & 2.50\sdec{5.00} & \textbf{84.08}\uinc{0.99} & \textbf{74.20}\udec{0.01} & \underline{52.30}\udec{0.15} & \textbf{5.6}\xinc{0.4} \\
SN-Tune & \cellcolor{orange!18}\textcolor{BurntOrange}{\textbf{SFT}} & 4.40\sdec{29.56} & 0.19\sdec{7.31} & 70.36\udec{12.73} & 68.01\udec{6.20} & 51.50\udec{0.95} & 42.0\xinc{36.8} \\
\textbf{SCA (Ours)} & \cellcolor{orange!18}\textcolor{BurntOrange}{\textbf{SFT}} & \underline{6.92}\sdec{27.04} & \underline{0.38}\sdec{7.12} & \underline{83.17}\uinc{0.08} & \underline{74.05}\udec{0.16} & 52.20\udec{0.25} & \underline{6.8}\xinc{1.6} \\
\cmidrule(lr){1-8}
LoRA    & \cellcolor{cyan!14}\textcolor{RoyalBlue}{\textbf{DPO}} & \textbf{0.00}\sdec{33.96} & \textbf{0.19}\sdec{7.31} & \textbf{86.81}\uinc{3.72} & 73.97\udec{0.24} & \textbf{53.10}\uinc{0.65} & 34.8\xinc{29.6} \\
Random       & \cellcolor{cyan!14}\textcolor{RoyalBlue}{\textbf{DPO}} & \textbf{0.00}\sdec{33.96} & \textbf{0.19}\sdec{7.31} & \underline{82.87}\udec{0.22} & \textbf{74.28}\uinc{0.07} & 52.20\udec{0.25} & \textbf{8.0}\xinc{2.8} \\
\textbf{SCA (Ours)} & \cellcolor{cyan!14}\textcolor{RoyalBlue}{\textbf{DPO}} & \textbf{0.00}\sdec{33.96} & \textbf{0.19}\sdec{7.31} & 81.73\udec{1.36} & \underline{74.21}\usame & \underline{52.60}\uinc{0.15} & \underline{17.6}\xinc{12.4} \\
\midrule
\rowcolor{violet!4}\multicolumn{8}{c}{\textbf{MISTRAL-7B-V0.1}} \\
Base Model         & --  & \basepadtwo{39.62} & \basepadtwo{30.00} & \basepadtwo{33.43} & \basepadtwo{53.74} & \basepadtwo{50.15} & \basepadone{0.8} \\
LoRA    & \cellcolor{orange!18}\textcolor{BurntOrange}{\textbf{SFT}} & \textbf{0.63}\sdec{38.99} & \textbf{0.58}\sdec{29.42} & 30.63\udec{2.80} & \underline{53.28}\udec{0.46} & \textbf{50.90}\uinc{0.75} & 7.6\xinc{6.8} \\
Random       & \cellcolor{orange!18}\textcolor{BurntOrange}{\textbf{SFT}} & 9.43\sdec{30.19} & \underline{5.19}\sdec{24.81} & \underline{31.01}\udec{2.42} & 52.95\udec{0.79} & \underline{50.45}\uinc{0.30} & \textbf{5.2}\xinc{4.4} \\
SN-Tune & \cellcolor{orange!18}\textcolor{BurntOrange}{\textbf{SFT}} & 5.03\sdec{34.59} & 5.77\sdec{24.23} & 28.81\udec{4.62} & 49.95\udec{3.79} & 49.15\udec{1.00} & 25.6\xinc{24.8} \\
\textbf{SCA (Ours)} & \cellcolor{orange!18}\textcolor{BurntOrange}{\textbf{SFT}} & \underline{2.52}\sdec{37.10} & \textbf{0.58}\sdec{29.42} & \textbf{32.68}\udec{0.75} & \textbf{53.75}\uinc{0.01} & 50.20\uinc{0.05} & \underline{5.6}\xinc{4.8} \\
\cmidrule(lr){1-8}
LoRA    & \cellcolor{cyan!14}\textcolor{RoyalBlue}{\textbf{DPO}} & \underline{16.35}\sdec{23.27} & \underline{5.58}\sdec{24.42} & \textbf{34.80}\uinc{1.37} & 53.60\udec{0.14} & \textbf{51.30}\uinc{1.15} & \underline{4.8}\xinc{4.0} \\
Random       & \cellcolor{cyan!14}\textcolor{RoyalBlue}{\textbf{DPO}} & 37.74\sdec{1.88} & 28.46\sdec{1.54} & \underline{34.12}\uinc{0.69} & \textbf{54.30}\uinc{0.56} & \underline{50.95}\uinc{0.80} & \textbf{2.0}\xinc{1.2} \\
\textbf{SCA (Ours)} & \cellcolor{cyan!14}\textcolor{RoyalBlue}{\textbf{DPO}} & \textbf{7.55}\sdec{32.07} & \textbf{4.23}\sdec{25.77} & 33.51\uinc{0.08} & \underline{53.82}\uinc{0.08} & 50.65\uinc{0.50} & 5.6\xinc{4.8} \\
\bottomrule
\end{tabular}
}
\vspace{-1.5em}
\end{table}

%% file: sections/6_conclusion.tex
\section{Conclusion}
\label{sec:conclusion}

In this work, we introduce SafeEvo, an interpretability framework that leverages differentiable mask optimization to attribute LLM refusal to sparse, functionally complete circuits and to trace their evolution across alignment checkpoints, moving beyond endpoint analyses of already-aligned models. Through extensive experiments on three LLMs, we reveal that weak refusal circuits already exist in pretrained-only models, causally govern their refusal of harmful queries, and are neither unique nor stable during safety post-training. We further bridge interpretability with alignment via SCA, showing that confining safety updates to pre-alignment refusal circuits yields stronger safety, less over-refusal, and better utility preservation. We believe SafeEvo offers a mechanistic account of how alignment reshapes LLMs, paving the way for more precise and interpretable alignment algorithms.

%% file: sections/7_ai_use_statement.tex
\section*{AI use statement}

We used a general-purpose language model as a writing aid, limited to polishing grammar and phrasing in prose the authors had drafted and tightening captions such as those of Figure 1 and Tables 2–4 to fit the page budget. We also used an AI coding assistant for routine chores: plotting boilerplate for the safety–utility trade-off curves in Figure 5 and the overlap heatmap in Figure 4, and formatting result dictionaries into the LaTeX of Tables 2, 3, and 9. AI tools played no role in generating research ideas, designing the extraction objective or the alignment method, running or interpreting experiments, or producing any reported number. All AI-assisted text and scripts were verified by the authors, who checked every rendered table and figure against the raw experiment logs.

\section*{Ethics statement}
This work studies the internal mechanisms of safety refusal in order to make alignment more effective and less costly to utility. All harmful prompts and completions come from established public safety benchmarks and alignment datasets used under their original licenses; we created no new harmful content, collected no data from human subjects, scored outputs automatically rather than exposing annotators to them, and release no harmful generations or attack artifacts. We acknowledge that our ablation experiments suppress refusal and are in principle an uncensoring recipe, but the intervention needs white-box weight access. More importantly, our work aims to understand the safety mechanisms in LLMs for better alignment methods.

\section*{Reproducibility statement}
We release anonymized code at the URL in the abstract; all models, datasets, and judges are public. The methods, baselines, and metrics are fully specified in the main text. The appendices then supply the settings and protocols behind the reported runs: Appendix A gives the circuit budgets, mask provenance, full extraction configuration, and our reimplementation of the external baseline; Appendix B the exact ablation values together with the intervention and refusal-scoring protocol and its controls; Appendix C the multi-seed and paired bootstrap procedure, so readers can tell which claims carry seed-level variance; Appendix D the XSTest scoring protocol; and Appendix E all training, generation, and evaluation hyperparameters.

%% file: sections/A_appendix.tex
\newpage
\section{Per-Model Sparsity and Mask Provenance}
\label{app:sparsity}

Table~\ref{tab:sparsity} reports authoritative trainable-parameter and
realized-circuit statistics from each circuit run's metadata. The trainable
fraction is tight across families ($0.35$--$0.40\%$), while the realized
active-neuron counts are model-specific. Table~\ref{tab:masks} identifies the masks used for ablation and SCA tuning; these masks are extracted on the untuned weights of the corresponding models.

\begin{table}[ht]
\centering\small
\caption{Per-model circuit budget (seed-1 representative; other seeds identical
recipe). Trainable params are fixed by LoRA rank $\times$ patched-Linear
dimensions, independent of mask size.}
\label{tab:sparsity}
\begin{tabular}{lrrrr}
\toprule
Family & active circuit neurons & patched Linears & trainable params & trainable \% \\
\midrule
Llama-3-8B  & 10{,}376 & 96 & 28.31M & 0.3513 \\
Qwen-2.5-7B & 3{,}594  & 84 & 30.28M & 0.3960 \\
Mistral-Inst-v0.1 & 5{,}724  & 96 & 28.31M & 0.3894 \\
\bottomrule
\end{tabular}
\end{table}

\subsection{External SN-Tune selector baseline}
\label{app:sn-baseline}

We reimplement the released PLND semantics of SN-Tune
\citep{zhao2025sntune}. On the official 200-prompt English harmful corpus, for
each prompt and layer we retain the top-1000 attention and top-2000 MLP
coordinates and intersect the prompt-level selections, capped at 100 per layer
and component. The resulting mask gates LoRA updates to \texttt{q}/\texttt{k}/
\texttt{v} output rows, MLP \texttt{up} output rows, and matching
\texttt{down} input columns. The selected and unselected gradients were audited
to be respectively nonzero and exactly zero.

\begin{table}[ht]
\centering\small
\setlength{\tabcolsep}{5pt}
\caption{SN-Tune selector sizes used in the external baseline. MLP coordinates
gate both \texttt{up} and \texttt{down}, so tuned projection slices equal
$q+k+v+2\,\mathrm{MLP}$.}
\label{tab:sn-counts}
\begin{tabular}{lrrrrr}
\toprule
Family & $q$ & $k$ & $v$ & MLP & tuned slices \\
\midrule
Llama-3-8B & 3{,}200 & 3{,}200 & 1{,}461 & 1{,}679 & 11{,}219 \\
Qwen-2.5-7B & 2{,}781 & 2{,}771 & 1{,}270 & 1{,}650 & 10{,}122 \\
Mistral-Inst-v0.1 & 3{,}200 & 3{,}200 & 1{,}107 & 1{,}548 & 10{,}603 \\
\bottomrule
\end{tabular}
\end{table}

\begin{table}[ht]
\centering\scriptsize
\setlength{\tabcolsep}{18pt}
\caption{\textbf{Mask provenance for ablation and SCA tuning.} Each row reports the circuit
object, neuron count, and extraction checkpoint used by that experiment.}
\label{tab:masks}
\begin{tabular}{llll}
\toprule
Experiment & Mask object & Neurons (Llama/Qwen/Mistral) & Extracted on \\
\midrule
Ablation (Section~\ref{sec:exp-exists}) &
$\mathcal{C}_1$ &
10{,}376 / 9{,}972 / 7{,}677 &
base weights \\
\sca{} tuning (Section~\ref{sec:exp-align}, Table~\ref{tab:main}) & $\mathcal{C}$ & 10{,}376 / 3{,}594 / 5{,}724 & base weights \\
\bottomrule
\end{tabular}
\end{table}

\subsection{Recorded extraction configuration}
\label{app:extraction-config}

Table~\ref{tab:extraction-config} records the final three starting-model
extractions used by the ablation in Section~\ref{sec:exp-exists}. These settings
instantiate Equation~\ref{eq:extraction-loss}; Table~\ref{tab:masks} separately
identifies the distinct mask object used by every experiment.

\begin{table}[ht]
\centering\scriptsize
\setlength{\tabcolsep}{5pt}
\caption{\textbf{Recorded circuit-extraction configuration.} Circuit-extraction configuration for the ablation experiments in Section 3.2.}
\label{tab:extraction-config}
\begin{tabular}{>{\raggedright\arraybackslash}p{0.22\linewidth}
                >{\raggedright\arraybackslash}p{0.70\linewidth}}
\toprule
Setting & Recorded value \\
\midrule
Data and split & LLM-LAT harmful-prompt records, each paired with a refusal and
harmful-compliance completion (4{,}948 available); first 100 for training and
the next 50 for validation; model-native chat template; maximum length 256. \\
Search space & One logit per output row of MLP \texttt{gate}/\texttt{up}/\texttt{down}; model weights
frozen; mask logits only are optimized in bf16. \\
Loss coefficients & $\alpha=1$, $\beta=1$, and
$\lambda_{\mathrm{mlp}}=0.05$, applied to
the sum of per-module mean soft masks in Equation~\ref{eq:extraction-loss}. \\
Mask relaxation & Logit initialization $q_0=0.2$; temperature $1$; hard forward
gate $\mathbf{1}[\sigma(q)>0.5]$ with a straight-through estimator. \\
Optimization & Fused AdamW, learning rate $10^{-2}$, batch size 4, no gradient
accumulation, linear schedule, zero warmup, zero weight decay. \\
Horizon and selection & 100-epoch schedule horizon; validation after each epoch;
checkpoint comparison uses the final validation minibatch's two completion
losses (sparsity excluded); stop after epoch 2 (50 optimizer updates), which was
the selected mask for all three families. \\
Extraction seeds & Llama-3-8B: 1; Qwen-2.5-7B: 2;
Mistral-7B-Instruct-v0.1: 2. \\
Deployed $C_1$ readout & Globally highest raw MLP
logits retained with $k=10{,}376/9{,}972/7{,}677$ for Llama/Qwen/Mistral. The
minimum selected logits were $0.496/0.471/0.496$, respectively. \\
Matched-random control & Prespecified seed 42; uniform without replacement
within each Linear, matching that Linear's deployed $C_1$ width. \\
\bottomrule
\end{tabular}
\end{table}

\begin{samepage}
\section{Ablation: extracted circuit versus a fixed matched-random mask}
\label{app:ablation}

Table~\ref{tab:ablation} reports the exact zero-ablation values behind
Figure~\ref{fig:ablation}.

\begin{center}
\begin{minipage}{\linewidth}
\centering\small
\captionsetup{hypcap=false}
\captionof{table}{\textbf{Starting-model circuit ablation.}
Explicit-refusal rate and raw LlamaGuard-3 ASR for the intact model,
$\mathcal{C}_1$ ablation, and a prespecified seed-42 width-matched random
ablation.}
\label{tab:ablation}
\begin{tabular}{llrrr}
\toprule
Family & Metric & base &
ablate $\mathcal{C}_1$ & ablate matched random \\
\midrule
Llama-3-8B  & refusal (\%) & 55.35 & \textbf{0.00} & 37.74 \\
            & ASR (\%)     & 48.43 & \textbf{91.19} & 49.06 \\
\midrule
Qwen-2.5-7B & refusal (\%) & 61.01 & \textbf{0.63} & 22.64 \\
            & ASR (\%)     & 32.70 & \textbf{88.68} & 64.15 \\
\midrule
Mistral-Inst-v0.1 & refusal (\%) & 35.85 & \textbf{0.63} & 42.14 \\
            & ASR (\%)     & 42.14 & \textbf{88.05} & 35.22 \\
\bottomrule
\end{tabular}
\end{minipage}
\end{center}
\end{samepage}

Refusal rate uses the classifier of \citet{xstest} applied to the first $150$
characters of each generation after truncation at the first conversational
boundary and removal of any leading verbatim prompt echo; ASR is raw
LlamaGuard-3 on those same outputs. For each
family, $\mathcal{C}_1$ is read from one extraction by retaining the globally
highest-scoring $10{,}376$/$9{,}972$/$7{,}677$ MLP entries for
Llama/Qwen/Mistral. The paired random control is a prespecified seed-42 draw
that matches the active count of $\mathcal{C}_1$ in every Linear. A null
(empty-mask) ablation
reproduces the base generation exactly, confirming that the intervention hook
is transparent. We use zero-ablation for both selected and random identities so
that intervention type and lesion width are held fixed.

For the 50-prompt sufficiency diagnostic in
Table~\ref{tab:circuit-sufficiency}, explicit refusals include both standard
refusal phrases and direct negative formulations that decline the harmful
request. Under this criterion, the intact Mistral model refuses 8 of 50
prompts (16.00\%).

\section{Multi-seed robustness and paired bootstrap}
\label{app:stats}

Table~\ref{tab:robustness} in the main text aggregates HarmBench raw ASR over
$n{=}4$ training seeds per condition (identical recipe, differing only in
training seed); The same table reports
the paired prompt-level bootstrap: for each family we resample the $159$ HarmBench
prompts with replacement and recompute the ASR difference (width-matched random
$-$ circuit) on each resample; all three $95\%$ intervals exclude zero. The
$n{=}18$ random ensemble on Llama (Section~\ref{sec:exp-align}) is disjoint
from the circuit seeds (worst circuit seed $8.81$ vs.\ best random $11.32$).

\section{XSTest coverage--over-refusal protocol}
\label{app:ood}

Table~\ref{tab:ood} in the main text jointly reports the two XSTest subsets
\citep{xstest}. The first value in each cell is raw Llama-Guard-3 ASR on the
$200$ unsafe-contrast prompts: harmful requests phrased to superficially
resemble the benign safe set. The second is the explicit-refusal rate on the
$250$ benign prompts, computed with the same truncated, de-echoed classifier
used for Table~\ref{tab:main}. Full-SFT obtains broader unsafe-contrast coverage
than circuit-SFT but over-refuses more; under DPO, circuit tuning reaches lower
unsafe-contrast ASR on Llama and Mistral, while Qwen full-DPO's near-total
reduction coincides with $34.8\%$ benign-prompt over-refusal.

\section{Hyperparameters}
\label{app:hparams}

Circuit and width-matched-random runs use LoRA rank $16$, $\alpha{=}16$,
dropout $0$, and bf16 on the MLP \texttt{gate}/\texttt{up}/\texttt{down}
projections with frozen attention; full-LoRA also patches all attention Linears.
SFT uses learning rate $10^{-4}$, $16$ epochs, maximum length $512$, and
effective batch $4$; the Llama circuit-SFT run uses batch $16$ for $4960$ steps
and a post-plateau checkpoint. DPO uses $\beta{=}0.1$, $3$ epochs, effective
batch $8$, and learning rate $10^{-4}$ for Llama/Qwen or $5\!\times\!10^{-5}$
for Mistral. Generation uses batch $16$, and Llama-Guard-3-8B judging uses
greedy decoding. The SN-Tune baseline matches the SFT data, rank, learning rate,
epochs, and maximum length, with batch $16$ for Llama and $4$ for Qwen/Mistral.
Utility uses the $5$-shot harness capped at $2000$ examples per task.